\documentclass{llncs}

\usepackage[utf8x]{inputenc}
\usepackage[pdftex]{graphicx}
\usepackage{algorithm}
\usepackage{algorithmic}
\usepackage{xspace}
\usepackage{rotating}

\begin{document}
\pagestyle{empty}  % switches off printing of running heads

%\title{Sign Language Gestures: can we Dissociate Meaning from Motion?
\title{Toward a Motor Theory of Sign Language Perception
}
%\titlerunning{Modeling Joint Synergies to Synthesize Realistic Movements}
\author{Sylvie Gibet, Pierre-Fran\c{c}ois Marteau and Kyle Duarte}

\institute{IRISA-UBS, Universit\'{e} de Bretagne Sud\\
\email{sylvie.gibet@univ-ubs.fr}
}

\maketitle

%%%%%%%%%%%%%%%%%%%%%%%%%%%%%%%%%%%%%%%%%%%%%%%%%%%%%%%%%%%%%%%%%%%%%%%%%%%%%%%%
%%%%%%%%%%%%%%%%%%%%%%%%%%%%%%%%%%%%%%%%%%%%%%%%%%%%%%%%%%%%%%%%%%%%%%%%%%%%%%%%
\begin{abstract}
Researches on signed languages still strongly dissociate linguistic issues related on phonological and phonetic aspects, and gesture studies for recognition and synthesis purposes. This paper focuses on the imbrication of motion and meaning for the analysis, synthesis and evaluation of sign language gestures. 
We discuss the relevance and interest of a motor theory of perception in sign language communication. According to this theory, we consider that  linguistic knowledge is mapped on sensory-motor processes, and propose a methodology based on the principle of a synthesis-by-analysis approach, guided by an evaluation process that aims to validate some hypothesis and concepts of this theory. Examples from existing studies illustrate the different concepts and provide avenues for future work. 

 \end{abstract}

%%%%%%%%%%%%%%%%%%%%%%%%%%%%%%%%%%%%%%%%%%%%%%%%%%%%%%%%%%%%%%%%%%%%%%%%%%%%%%%%
\section{Introduction}
%%%%%%%%%%%%%%%%%%%%%%%%%%%%%%%%%%%%%%%%%%%%%%%%%%%%%%%%%%%%%%%%%%%%%%%%%%%%%%%%
The ever growing use of gestures in advanced technologies, such as augmented or virtual reality environments, requires more and more understanding of the different levels of representation of gestures, from meanings to motion characterized by causal physical and biological phenomena. This is even more true for skilled and expressive gestures, or for communicative gestures such as sign language gestures, both involving high level semiotic and cognitive representations, and requiring extreme rapidity, accuracy, and physical engagement with the environment. 

In this paper we highlight the tight connection between the high-level and low-level representations involved in signed languages. Signed languages are fully-formed languages that serve as a means of communication between deaf people. They are characterized by meanings: they have their own rules of compositional elements, grammatical structure, and prosody; but they also include multimodal components that are put into action by movements. 
They are indeed by essence multi-channel, in the sense that several modalities are implicated when performing motion: body, hands, facial expression, gaze direction, acting independently but participating all together to convey meaningful and discriminative information. In signed language storytelling for example, facial expressions may be used to qualify actions, emotions, and enhance meaning.

We focus on data-driven models, which are based on observations of real signed language gestures, using captured motion or videos. Motion capture allows us to find relevant representations that encode the main spatio-temporal characteristics of gestures. In the same way, analyzing videos may lead to annotations where significant labels indicate the morpho-syntactic nature of elements composing gestures, and may constitute a starting point for determining phonetic structures. By combining both pieces of information, motion capture data and videos, we may also extract accurate low and high level features that help to understand sign language gestures. We believe that data-driven methods, incorporating constraints extracted from observations, significantly improve the quality and the credibility of the synthesized motion. To go beyond, we propose this synthesis-by-analysis method, corrected by a perceptual evaluation loop, to model the underlying mechanisms of signed language gesture production.\\

In the remainder of the paper, we propose a guideline aiming at characterizing the role of sensory-motor information for signed language understanding and production, based on the motor theory of sign language perception. We then provide a general methodology for analyzing, synthesizing, and evaluating signed language gestures, where different sensory data are used to extract linguistic features and infer motor programs, and to determine the action to perform in a global action-perception loop. The different concepts and models are illustrated by related works, both from the points of view of signed language linguistics and movement science communities.

After describing related works in the next section, we propose sign language production and perception models underlying the motor theory of sign language perception. A methodology is then proposed to highlight how this theory may be exploited in both theoretical sign language research and motion sciences.

%%%%%%%%%%%%%%%%%%%%%%%%%%%%%%%%%%%%%%%%%%%%%%%%%%%%%%%%%%%%%%%%%%%%%%%%%%%%%%%%
\section{Related Works}
%%%%%%%%%%%%%%%%%%%%%%%%%%%%%%%%%%%%%%%%%%%%%%%%%%%%%%%%%%%%%%%%%%%%%%%%%%%%%%%%

There are two main approaches in modeling and producing sign language gestures, that are addressed differently in the different research communities: the first one, addressed by the signed language linguists, concerns the formation of the meaning from observations; the second one, addressed by motion science researchers, is related to motion generation and recognition from high-level sign descriptions. Most of the time, these two approaches are considered separately, as the two research communities do not share the same tools and methods. \\

Linguistic researchers work on signed languages from some observation of natural utterances, most often through video data: they build theories describing the mapping between these observations and linguistic components (phonetics, phonological structures, etc.).
The resulting models are still widely debated in the sign language community, and usually, motion characterization is not seen as a prime objective for elaborating phonological model \cite{Hulst2001} or phonetic model \cite{JohnsonLiddell10}.  In order to validate their observations and analysis, they need better knowledge of movement properties: kinematic invariants within signs and between signs, physical constraints, etc. Invariant laws in movements are discussed in \cite{Gibet03}. \\

Movement researchers on the other hand (bio-mechanicians, neuroscientists, computer animators, or roboticians) try to build simulation models that imitate real movements. Their approach consists, from high-level descriptions (planning), of specifying a sequence of actions as a procedural program.
They need to acquire better knowledge of the rules governing the system behavior, such as syntactic rules or parameterization of the sign components according to the discourse context. The next problem consists of interpreting these rules using specific computer languages (from scripting languages to procedural or reactive languages), and traducing them into sensory-motor processes underlying the physical system that produce movement.

Most of the works in this area focus on the expressivity of the high-level computer languages, using descriptive or procedural languages, for example the XML-based specification language called SiGML \cite{Kennaway2003} which is connected to  the HamNoSys \cite{Prillwitz1989} notation system, and interpreted into signed language gestures using classical animation techniques. A more exhaustive overview of existing systems using virtual signers technology can be found in \cite{Gibet11}. For these kinds of applications involving signed language analysis, recognition, translation, and generation, the nature of the performed gestures themselves is particularly challenging. 

Alternatively, data-driven animation methods can be substituted for these pure synthesis methods. In this case the motions of a real signer are captured  with different combinations of motion capture techniques. 
Though these methods significantly improve the quality and credibility of animations, there are nonetheless several challenges to the  reuse of motion capture data in the production of sign languages. Some of them  
are related to the spatialization of the content, but also to the rapidity and precision required in motion performances, and to the dynamic aspects of movements. All these factors are responsible for phonological inflection processes. Incorrectly manipulated, they may lead to imperfections in the performed signs (problems in timing variations or synchronization between channels) that can alter the semantic content of the sentence.
A detailed discussion on the important factors for the design of virtual signers in regard to the animation problems is proposed in  \cite{Courty10}. \\

Little has been done so far to determine the role of sensory-motor activity for the understanding (perception and production) of signed languages. The idea that semantic knowledge is embodied into sensory-motor systems has given rise to many studies, bringing together researchers from domains as different as cognitive neuroscience and linguistics, but most of these works concern spoken languages. 
This interaction between language and action are based on different claims such as:
\begin{itemize}
\item{imagining and acting share the same neural substrate \cite{Goldman06}};
\item{language makes use in large part of brain structures akin to those used to support perception and action \cite{Arbib08}.}
\end{itemize}

Among these recent research interests, some researchers share the idea that motor production is necessarily involved in the recognition of sensory (audio, visual, etc.) encoded actions; this idea echoes what is called the \emph{motor theory of speech perception} which holds that the listener recognizes speech by activating the motor programs that would produce sounds like those that are being heard \cite{Liberman67}. Within this theory, sensory data are auditory or visual clues (mouth opening), and the motor actions are vocal gestures (movements of the vocal tract, tongue, lips, etc.).

This theory can be easily transposed to sign languages, and we will call it the \emph{Motor Theory of Sign Language Perception}. In this case too, the linguistic information is embodied into sensory-motor processes, where sensory data may be visual clues (iconic gestures, classifiers), or perception of action (contact between several body parts, velocity or acceleration characteristics, etc.).

\section{The Motor Theory of Sign Language Perception}
%%%%%%%%%%%%%%%%%%%%%%%%%%%%%%%%%%%%%%%%%%%%%%%%%%%%%%%%%%%%%%%%%%%%%%%%%%%%%%%%

All the evidence briefly reported in the previous section tends to show that perception and production of language utterances are closely related. It remains to describe or model this relationship. At the light of this evidence, the motor theory of speech perception, which states that what we perceive is nothing but the movement of the articulatory system (body movements), suggests that part of conceptual and language structures are encoded at motor program levels, e.g. as a sequence of motor actions allowing to produce the desired sensory (or perceptive) effect. 

Similarly to the motor theory of speech perception, the motor theory of sign language perception that we promote in this paper claims that what we perceive is the movement of body articulators, and that the encoding and decoding of linguistic information should be partly addressed at motor program level characterizing the movement intention. Furthermore, if we accept the idea that the motor program level  is where the linguistic cues are encoded, then the motor theory of perception leads to consider that we can infer motor programs from observed sensory cues only (motor act). We call this inference an inversion process since its purpose is to deduce the cause from the consequence (sensory observation).\\

Therefore, if we go further in the modeling of these concepts, we assume that the motor theory of sign language perception is based on two inversion mechanisms, one for sign language production, and the other one for sign language perception. These mechanisms will be used as part of encoding and decoding processes of linguistic units. By linguistic units we mean here phonetic and phonological elements specific to sign languages.\\

\begin{figure}[tb]
\centering{}
\includegraphics[width=0.8\textwidth]{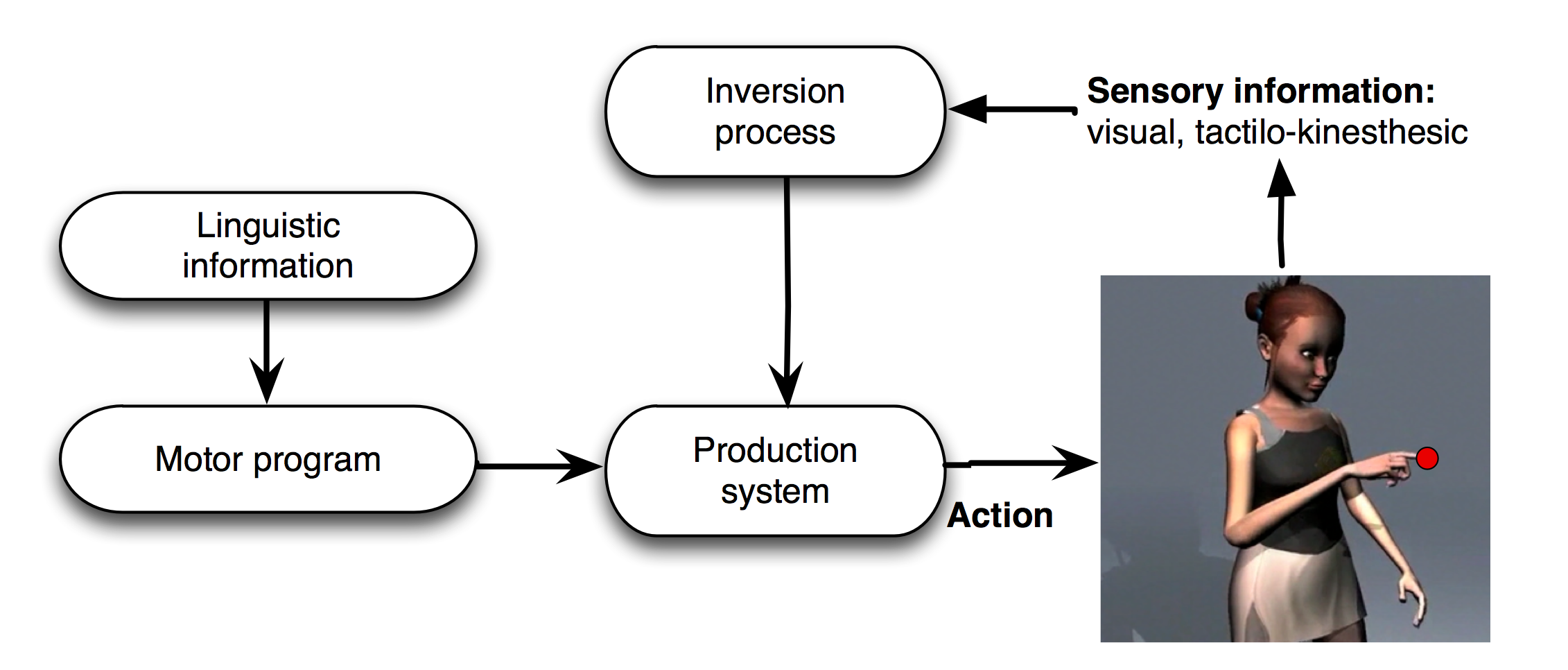}
\caption{Sign language production: encoding from motor program and linguistic information}
\label{fig:sensorymotor2}
\end{figure}

The first inversion process for sign language production, also called encoding process, is represented in Figure~\ref{fig:sensorymotor2}. It is a closed-loop system, where the signer uses sensory information to produce the desired actions corresponding to a specific motor program. The signer performing gestures perceives the environment through many sensory cues: he can view his interlocutor, and also the entities positioned in the signing space (spatial targets); he may also capture auditive, tactile (perception of touch), proprioceptive (perception of muscles and articulations), and kinesthetic clues (perception of velocity, acceleration, etc.) from its own body movements. These sensory cues are then inverted to provide motor commands that modify the current action applied to the musculor-skeleton system. When producing sign language gestures, the linguistic information is also exploited to generate a sign language utterance which is then translated into a motor program. 

In the context of sign language synthesis, the motor programs may be represented by a sequence of goals, as for example key postures of the hand, or targets in hand motion, or facial expression targets. These targets are then interpreted into continuous motion, through an inverse kinematics or dynamics solver \cite{Gibet1994}, \cite{Lebourque99}, \cite{GibetM03}.\\

\begin{figure}[tb]
\centering{}
\includegraphics[width=0.8\textwidth]{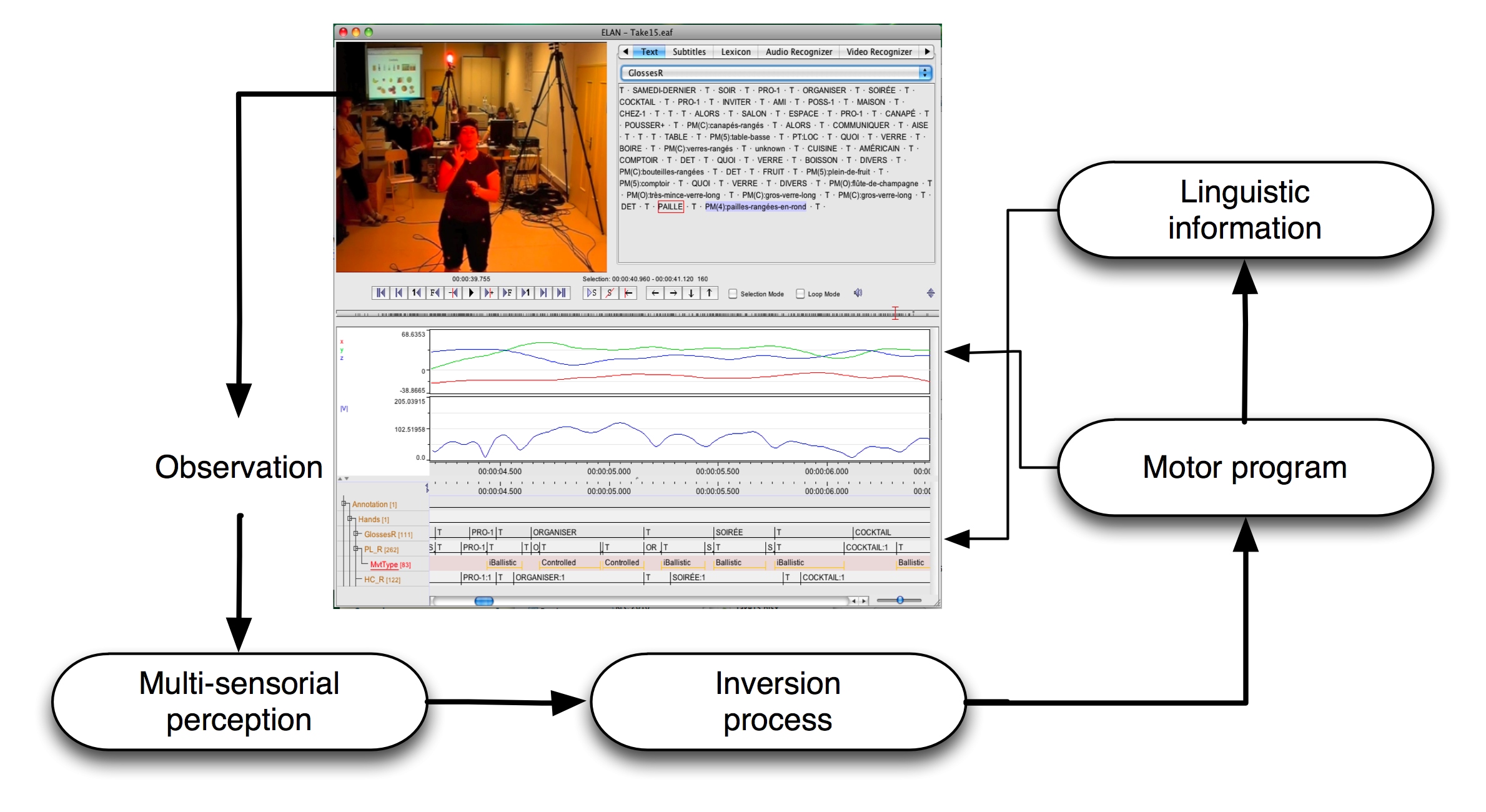}
\caption{Sign language perception: decoding for inferring motor program and extracting linguistic information}
\label{fig:sensorymotor3}
\end{figure}

The second inversion process used for gesture perception, also called decoding process, is represented in Figure~\ref{fig:sensorymotor3}. From the observation of a signer, it consists in extracting multi-sensorial cues, and then to simultaneously infer motor programs (allowing to reproduce the detected sensory cues), and extract linguistic information.

Our approach to sign language perception can be divided into two kinds of analysis studies. The first one consists of a linguistic analysis that tries to extract phonetic or phonological features from the observation of signers. The second one consists in finding invariants or motor schemes in the data, above which one can build linguistic knowledge.

This last approach, inspired from the neuroscience community, may exploit statistical tools in order to extract some regular features or schemes embodied to motion data.

%%%%%%%%%%%%%%%%%%%%%%%%%%%%%%%%%%%%%%%%%%%%%%%%%%%%%%%%%%%%%%%%%%%%%%%%%%%%%%%%
\section{Methodology: Sign Language Production and Perception}
%%%%%%%%%%%%%%%%%%%%%%%%%%%%%%%%%%%%%%%%%%%%%%%%%%%%%%%%%%%%%%%%%%%%%%%%%%%%%%%%

\begin{figure}[tb]
\centering{}
\includegraphics[width=1.0\textwidth]{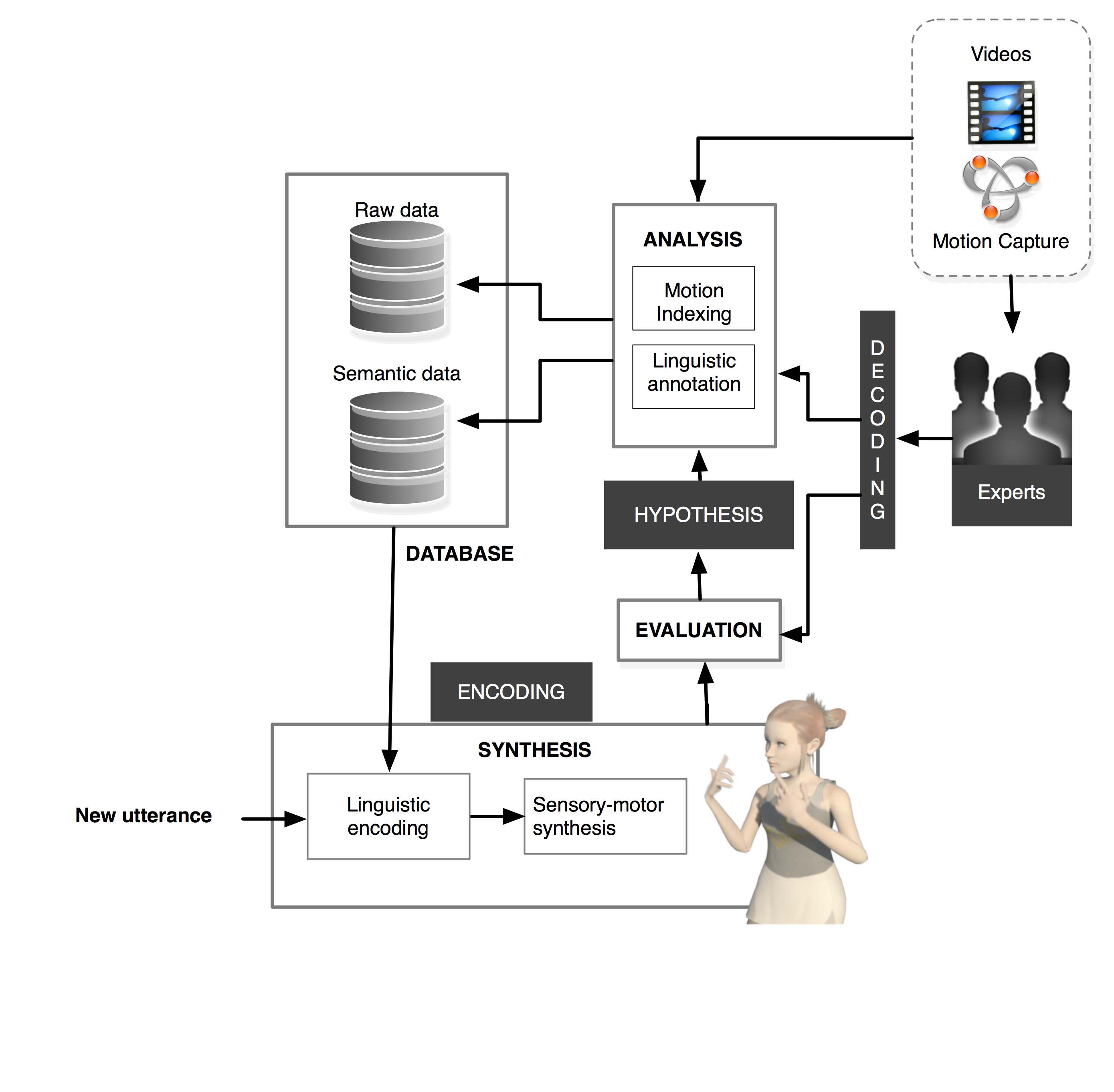}
\caption{Analysis, synthesis, and evaluation methodology}
\label{fig:methodology}
\end{figure}

In practice, production and perception are closely linked in a language communication context. In order to study jointly both mechanisms, we propose a general and experimental methodology based on an analysis (perception) / synthesis (production) approach, depicted in Figure~\ref{fig:methodology}. It contains the following three building blocks.

\begin{itemize}
\item{i) The Analysis block refers to the perception or decoding aspect of the methodology. It uses observed information from simultaneously captured motion data and videos. It is based on hypothesis related to the linguistically encoded structure of signs, and the motor programs underlying the performed gestures.
In practice, given the different nature of information that should be encoded (symbolic and numerical), it is more efficient to process and store data in two different structures, namely a semantic database for linguistic annotations, and a raw database for motion capture data;}

\item{ii) The Synthesis block covers the production or encoding aspect of the methodology. It is composed of a sensory-motor animation system which uses both a scripting process expressing a new utterance and  the corresponding motor program that uses pre-recorded motion chunks. Moreover, a 3D rendering engine allows to visualize the virtual signer performing the signs;}

\item{iii) The Evaluation block makes possible the evaluation of the analysis hypothesis, at the light of the synthesized gestures. Deaf experts or sign language signers may indeed qualify the different performances (quality of the gestures, realism, understandability), and propose some changes of the models and sub-segment structures including motor program schemes. We conjecture that during evaluation, based on their own sensory-motor inversion loop, experts or signers are implicitly able to validate or invalidate the synthesized motor performance and subsequently the hypothesis that have been made for the elaboration of the motor programs.}\\
\end{itemize}

This analysis-by-synthesis methodology requires to bring together researchers from different communities.
Preliminary work has been undertaken on the basis of the collected data within the project \emph{SignCom} \cite{signcom}. Some models and results underlying the former methodology are presented below, in the context of analysis of French sign language corpora, and data-driven synthesis. The use of  3D avatars driven by semantic and raw motion databases also allows us to go further the restrictions of videos, and to evaluate the feasibility and understandability of the models. 

\paragraph{\bf Corpus and database} 

%A case study of such a system can be represented by the project \emph{SignCom}. This project aims to improve the quality of real-time interaction between humans and virtual agents by exploiting French Sign Language gestures. A global overview of this project is shown in and presented at URL \cite{signcom}.

%Designing the corpus, we first had a reflexion on the linguistic categorization, and included into the corpus:
%\begin{itemize}
%\item{lexical items of various kinds and motion forms;}
%\item{non-lexical signs, the so-called iconic gestures or classifiers: for these signs, we mapped different sizes and shapes (examples of the sign PLATE, or GLASSWARE).}
%\end{itemize}
%
%We also included many instances of directed verbs and pointing gestures, such as: LIKE vs. NOT-LIKE, GIVE 
%vs. TAKE, 1-INVITE-x (I invite x) vs. x-INVITE-1 (x invites me), PUT, TAKE OFF, etc. (see Figure \ref{fig:like}).
%
%\begin{figure}[tb]
%\centering{}
%\includegraphics[width=0.8\textwidth]{Figures/LIKE-LIKE-NOT}
%\caption{Representation of the sign "LIKE / LIKE-NOT", and the sign "GIVE / TAKE"; the movements of the signs are inverted}
%\label{fig:like}
%\end{figure}

 The observational data are composed of 50 minutes of sign language motion captured data which gather data recorded with 43 body markers, 41 facial markers, and 12 hand markers, and videos of the same sequences recorded with one camera. Some of the challenges posed by the corpus creation and the capture of heterogenous data flows are detailed in \cite{DuarteGibet2010b} and \cite{DuarteGibet2010}. It should be noted that the choice of the corpus (choice of the thematics, limited vocabulary, choice of lexical and non-lexical signs, motion forms, etc.) may potentially influence the analysis and synthesis processes. 
 
% **We included in our corpus many instances of directed verbs and pointing gestures, such as: LIKE vs. NOT-LIKE, GIVE vs. TAKE, 1-INVITE-x (I invite x) vs. x-INVITE-1 (x invites me), PUT, TAKE-OFF, etc. (see Figure \ref{fig:like}).**
%
%\begin{figure}[tb]
%\centering{}
%\includegraphics[width=0.8\textwidth]{Figures/LIKE-LIKE-NOT}
%\caption{Representation of the sign "LIKE / LIKE-NOT", and the sign "GIVE / TAKE"; the movements of the signs are inverted}
%\label{fig:like}
%\end{figure}
%

\paragraph{\bf Analysis} The previous corpus has been analyzed and indexed by sign language experts: we separated the linguistic indexing from the raw motion indexing.

 \begin{itemize}
\item {\bf The linguistic indexing} is provided by annotations performed by sign language linguists associated to deaf people. 
Signs are generally decomposed into various components, such as location, handshape, and movement as proposed by Stokoe \cite{Stokoe1960}.  Since then, other linguists have expanded and modified Stokoe's decompositional system, introducing wrist orientation, syllabic patterning, etc. \cite{JohnsonLiddell10}. However, signed languages are not restricted to conveying meaning via the configuration and motion of the hand, but instead involve the simultaneous use of both manual and non-manual components. The manual components of signed language include hand configuration, orientation, and placement or movement, expressed in the signing space (the physical three-dimensional space in which the signs are performed). Non-manual components consist of the posture of the upper torso, head orientation, facial expression, and gaze direction.\\

Following this structural description of signs, we annotate the selected corpus, identifying each sign type found in the video data with a unique gloss so that each token of a single type can be easily compared, and segmenting the different tiers composing the signs by exploiting grammatical and phonological models \cite{JohnsonLiddell10}. The structure of the annotation scheme is characterized by: 
\begin{itemize}
\item{a spatial structure, defined by several tiers and a structural organization by gathering several channels;}
\item{a temporal structure, resulting from manual and semi-automatic segmentation, allowing transitions / strokes labelling;}
\item{a manual labeling with elements and patterns borrowed from linguists; we have followed the phonetics model of Johnson and Liddell \cite{JohnsonLiddell10}.\\
}
\end{itemize}

This annotation scheme allows to match motion data and phonetic structure, as 
shown in figure \ref{fig:annotation}, thus providing ways to index synchronously the motion to the 
phonetic tiers.\\

\begin{figure}[tb]
\centering{}
\includegraphics[width=1.0\textwidth]{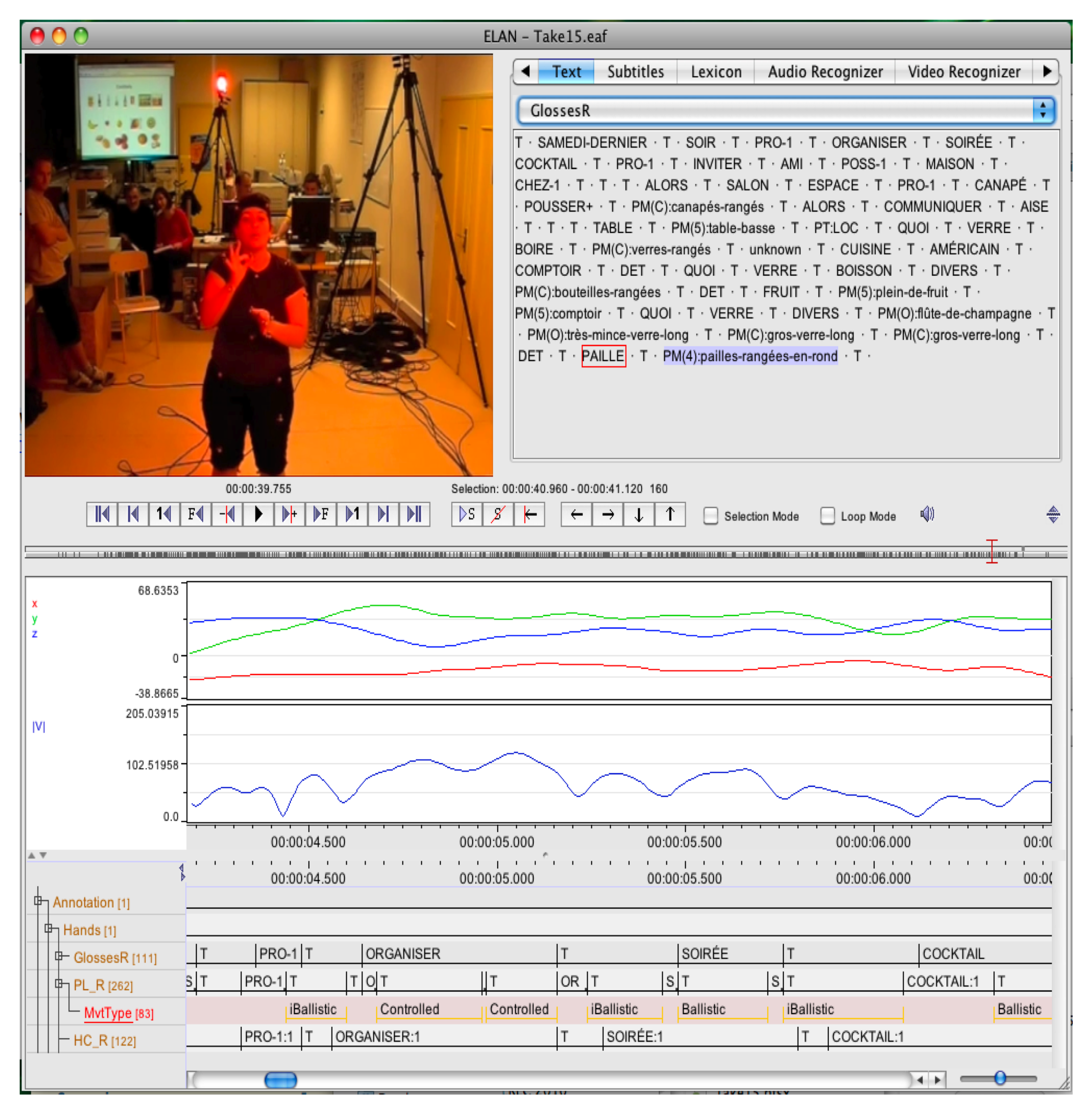}
\caption{A proposed annotation scheme, matching motion data and phonetic structure}
\label{fig:annotation}
\end{figure}

%It should be noted however that the phonetic modeling of signs is still an open question, thus structuring signs both spatially and temporally may sometimes fail in forming new signs or utterances.\\

\item {\bf The motion indexing} is based on motion processing.
Sign language data have already been studied, following different approaches. We first identified phonological items, described as sequences of motion targets and handshape targets  \cite{Lebourque99}, and used motor control rules, as the ones described in \cite{Gibet03}, to produce realistic hand motion.

Using motion captured data from French sign language corpora, we have also developed specific analysis methods that have led to the extraction of low-level or high-level motor schemes. We first automatically segmented handshape sequences \cite{Heloir05}, or hand movements that may be correlated to motor control laws \cite{Gibet08}. Secondly, statistical analysis have been conducted to characterize the phasing between hand motion and handshape \cite{Heloir07}, or to categorize hand motion velocity profiles within signs or during transitions between signs \cite{Duarte2010} (controlled, ballistic, and inverse-ballistic movements).
Similar works have been carried out to show the temporal organization in Cued Speech production \cite{AttinaBCO04}.

%\begin{itemize}
%%\item{Data reduction methods \cite{Heloir06};}
%\item{Invariant extracting, such as motor control rules \cite{Lebourque99}. In this work we have identified phonological items, described as sequences of motion targets and handshape targets;}
%\item{Segmenting the data automatically \cite{Heloir05}, \cite{Gibet08};}
%\item{Determining statistical properties \cite{Heloir07}, \cite{Duarte2010}.}
%\end{itemize}

\end{itemize}

We also implemented a two-levels indexed database (semantic and raw data) \cite{Awad09}. From such database, it will be possible to go further in the statistical analysis, and thus extract other invariants features and motor schemes, and to use them for re-synthesis. \\

\paragraph{\bf Synthesis} Conversely, using these enriched databases to produce new utterances from the corpus data remains challenging regarding the hypothesis derived from the analysis processes. Different factors may be encoded into the motor program driving the synthesis engine, such as the dynamics of the gestures (velocity profiles, etc.), the synchronization between the channels, or the coarticulation effects by using the sequence of pre-identified targets.

The multichannel animation system for producing utterances signed in French Sign Language (LSF) by a virtual character is detailed in \cite{Gibet11}.  \\

\paragraph{\bf Evaluation} 
Concerning evaluation issues, the idea is not so much to evaluate the signing avatar, but to evaluate the different hypotheses related to the decoding of signs, from the observation of sign language performances, and to the corresponding encoding of signs within the synthesis system. With this analysis-by-synthesis approach it is possible to possibly refine the different hypothesis and to help understanding the coupled production-perception mechanisms.

%Concerning evaluation issues, the idea is not so much to evaluate the signing avatar, but to evaluate the different hypotheses related to the decomposition of signs and utterances into significant elements, and the possibility of recombining these elements to produce new utterances. 

Currently, the research community focuses on the usability of the avatar. The evaluation process can be divided into two processes: i) the evaluation of the acceptability of the avatar, which can be measured by human-likeness, coordination, fluidity, realism of the three-dimensional rendering; ii) the evaluation of the understandability of the avatar, which requires the recognition of signs by measuring the precision of the signs, the co-articulation effects, etc., and measuring the level of recognition of the sentences and the story. A preliminary evaluation has been performed in \cite{Gibet11}. Understanding, characterizing more thoroughly the production and perception of sign language in the context of a motor theory of perception is a natural and promising perspective that should be carried out in the near future.

%%%%%%%%%%%%%%%%%%%%%%%%%%%%%%%%%%%%%%%%%%%%%%%%%%%%%%%%%%%%%%%%%%%%%%%%%%%%%%%%
\section{Conclusion}
%%%%%%%%%%%%%%%%%%%%%%%%%%%%%%%%%%%%%%%%%%%%%%%%%%%%%%%%%%%%%%%%%%%%%%%%%%%%%%%%
This paper promotes a motor theory of sign language perception as a guideline for the understanding of linguistic encoding and decoding of sign language gestures. According to this theory, what we perceive is nothing but the movement of the body's articulators. In other words, this assumption states  that the linguistic knowledge is mapped onto sensory-motor processes. Such an a priori statement relies on two main hypothesis: firstly, we are able to infer motor data from sensory data through a sensory motor inversion process, and secondly, elements of linguistic information are somehow encoded into motor programs. A methodology straightforwardly derived  from these two hypothesis and based on a so-called analysis-by-synthesis loop is detailed. This loop, through a perceptive evaluation carried out by sign language experts, allows to validate or invalidate hypothesis on linguistic encoding at motor program levels.
Although much work remains to be done to validate the methodology and the motor theory of sign language perception itself, its feasibility and practicality has been demonstrated in the context of French sign language corpora analysis and data-driven synthesis. 

It should be noted that the study of sign languages is a favorable field for validating motor theories of perception, since it is rather easy to infer the articulators' movements from sensory data (captured data and videos). However, this promising interdisciplinary research orientation requires the involvement of sign language linguists, deaf signers, neuroscientists and computer scientists.

\subsection*{Acknowledgments}
This work has been partially supported by the $SignCom$ project, an Audiovisual and Multimedia project by the French National Research Agency (ANR). 

%%%%%%%%%%%%%%%%%%%%%%%%%%%%%%%%%%%%%%%%%%%%%%%%%%%%%%%%%%%%%%%%%%%%%%%%%%%%%%%%
\bibliographystyle{unsrt}
\bibliography{biblio}
%%%%%%%%%%%%%%%%%%%%%%%%%%%%%%%%%%%%%%%%%%%%%%%%%%%%%%%%%%%%%%%%%%%%%%%%%%%%%%%%

%%%%%%%%%%%%%%%%%%%%%%%%%%%%%%%%%%%%%%%%%%%%%%%%%%%%%%%%%%%%%%%%%%%%%%%%%%%%%%%%
\end{document}